\documentclass[aps,floats,floatfix,twocolumn]{revtex4-2}

\usepackage{amsmath,amssymb,bm,graphicx,amsfonts,booktabs}
\usepackage{array}[=2016-10-06]
\usepackage{multirow,tabularx}
\usepackage{enumitem}
\usepackage{xcolor}
\usepackage[section]{placeins}
\usepackage{hyperref}
\hypersetup{
  colorlinks=true,
  linkcolor=blue!50!black,
  citecolor=blue!50!black,
  urlcolor=blue!50!black
}

\newcommand{\ndcg}[1][10]{\mathrm{nDCG\text{@}#1}}

\graphicspath{{plots/}}

\begin{document}

\title{Bias Robustness of Genetic Multi-Criteria Sorting versus Learned Rankers under Proxy Skew and Label Contamination}

\author{Vladyslav Shevchuk$^{1}$\thanks{Corresponding author. URL: https://danube.ai}}
\author{Carlos Uziel P\'erez~Malla$^{2}$}
\author{Philipp Wissgott$^{1}$}
\affiliation{$^1$ danube.ai solutions GmbH, 1040 Vienna, Austria}
\affiliation{$^2$ mypaperwork.ai, 1020 Vienna, Austria}

\begin{abstract}
Learning-to-rank (LTR) systems often absorb undesirable signals that leak into training labels or appear as cheap structural proxies among features. We study whether \emph{label-free} genetic multi-criteria sorting, which reranks fixed candidate pools by an explicit criteria vector~$C$, occupies a more favourable bias--utility operating point than label-trained LTR. We run two listwise studies on the Synthetic, MSLR-WEB10K, Baidu-ULTR, WANDS, and Expedia datasets. The proxy-skew study keeps labels clean but exposes a calibrated proxy~$P$: Genetic~AI algorithms show substantially lower rank--proxy correlation (Spearman correlation between ranks and~$P$) than LightGBM, XGBoost, and RankSVM (e.g.\ $\approx 0.15$--$0.16$ vs.\ $0.59$--$0.87$ on WANDS). A supporting label-contamination study contaminates training labels with a contaminant~$B$; Genetic~AI algorithms are label-free and thus remain unaffected by label contamination (bias curves stay flat), while LightGBM and XGBoost bias curves rise (OLS slope of rank--$B$ correlation vs.\ contamination, $\texttt{bias\_slope}$, reaches $0.58$--$0.65$ on WANDS/Baidu). Every bias claim is paired with clean-label $\ndcg$ on a bias--utility frontier. Genetic sorting is best read as an architectural alternative that optimizes declared attributes on a fixed pool, rather than as a universal replacement for LTR on raw relevance.
\end{abstract}


\maketitle

\onecolumngrid
\begin{center}
\includegraphics[width=\textwidth]{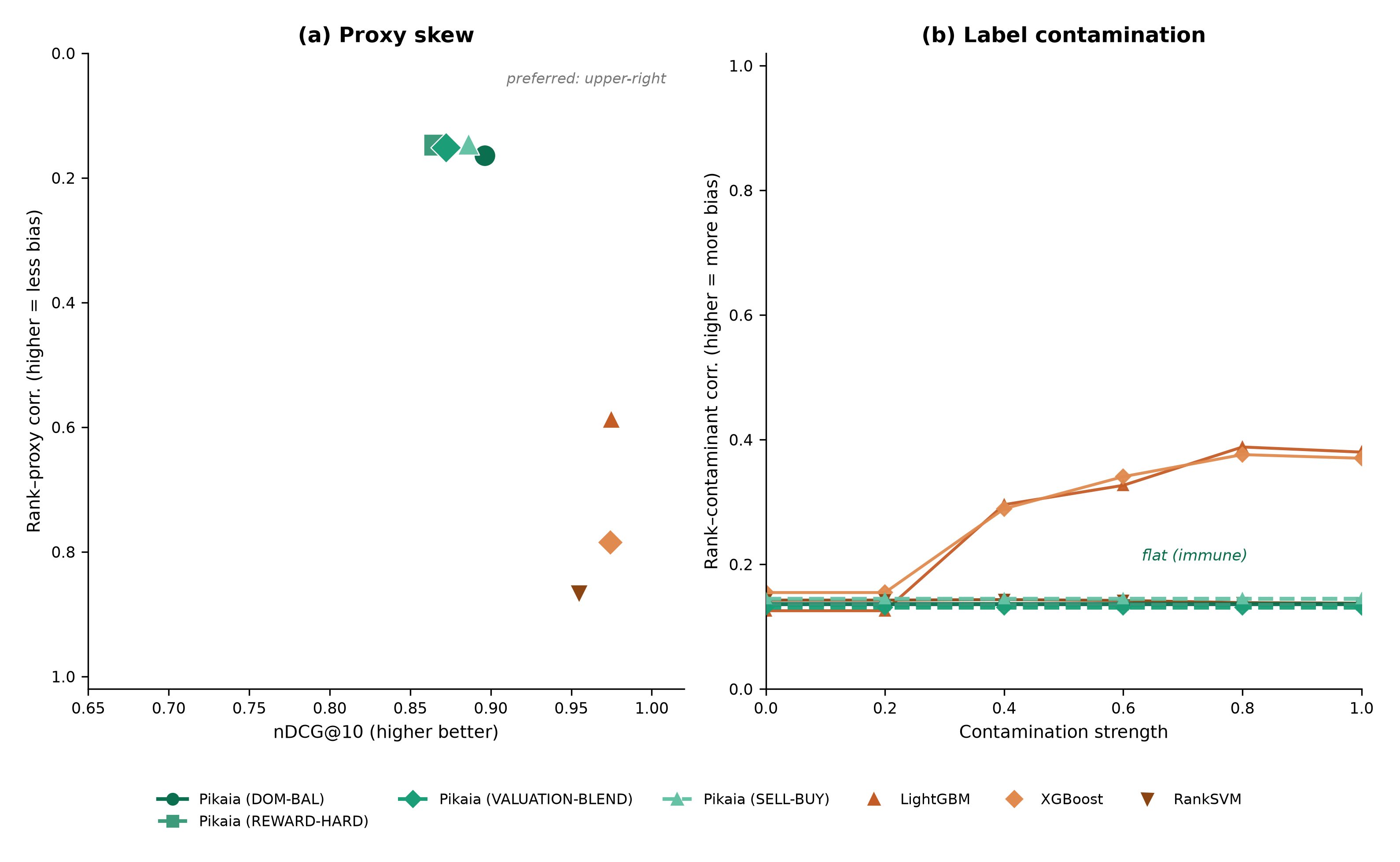}\\[0.6em]
\refstepcounter{figure}
\textbf{\figurename~\thefigure.}~
(a)~Proxy skew: clean-label $\ndcg$ against rank--proxy correlation (Spearman correlation between ranks and proxy~$P$). \texttt{Pikaia} strategies occupy the preferred upper-right region (high utility with low proxy alignment), while LightGBM, XGBoost, and RankSVM sit lower with stronger proxy alignment.
(b)~Label contamination: rank--contaminant correlation (Spearman correlation between ranks and undesignated signal~$B$) as training labels are increasingly mixed with~$B$. Open \texttt{pikaia} strategies stay flat because they never refit on contaminated labels; LTR absorbs~$B$ as contamination rises.
Methods: \emph{Pikaia~(DOM-BAL)}, \emph{Pikaia~(REWARD-HARD)},
\emph{Pikaia~(VALUATION-BLEND)}, and \emph{Pikaia~(SELL-BUY)} are open Pikaia strategy configurations;
\emph{LightGBM}, \emph{XGBoost}, and \emph{RankSVM} are the learned LTR peers.
\label{fig:visual_abstract}
\end{center}
\vspace{0.8em}

\clearpage
\twocolumngrid

\section{Introduction}
\label{sec:intro}

Ranking pipelines fail in two common ways that are easy to miss if evaluation tracks only retrieval utility. First, \emph{proxy skew}: even with clean labels, tree and linear models overweight cheap structural correlates of relevance (document length, review count, brand flags, click counts) instead of the attributes the task is meant to optimize~\cite{geirhos2020shortcut,chen2023bias,abdollahpouri2019unfairness}. Second, \emph{label contamination}: popularity, clicks, or presentation position leak into training grades, and label-trained LTR absorbs the contaminant because it is rewarded for predicting contaminated labels~\cite{joachims2017unbiased,joachims2007evaluating}.

Genetic multi-criteria sorting algorithms (open \texttt{pikaia} strategy variants) do not train on relevance labels. They rerank a fixed candidate pool by an explicit criteria vector~$C$ (attributes, polarity, optional weights), following the Genetic~AI / Evolutionary Data Theory line of work in which candidates and features are treated as organisms and genes in an evolutionary game~\cite{wissgott2025geneticai,wissgott2026edt,daniilidis2026featureweighting}. That architectural difference is the scientific object of this work: \emph{does refusing to fit proxy features or biased labels buy a better bias--utility trade-off?}

Prior work on bias in ranking spans unbiased LTR under position bias~\cite{joachims2017unbiased,agarwal2019estimating,oosterhuis2023doubly}, societal bias in neural rankers~\cite{rekabsaz2020neural,rekabsaz2021societal}, and fair exposure allocation~\cite{singh2018fairness,zehlike2022fair,ekstrand2022fairness}. Evolutionary methods have long been used to discover ranking functions~\cite{fan2004generic,tahery2021ga}, but they are rarely evaluated as \emph{bias-robust} alternatives to label-trained LTR under controlled proxy and contamination protocols. We fill that gap with a two-study design that shares candidate lists and metrics but separates failure modes, with proxy refusal as the main claim.

\paragraph{Contributions.}
\begin{enumerate}[leftmargin=1.5em,itemsep=0.2em]
  \item Proxy-skew study: clean-label proxy experiments showing default Genetic~AI algorithm configurations remain far less proxy-aligned than LightGBM, XGBoost, and RankSVM, with ablation utility loss when~$P$ is removed reported for Genetic~AI algorithms vs.\ LightGBM on key domains.
  \item Label-contamination study (supporting): contamination sweeps that confirm the architectural claim. Genetic~AI algorithm rank--$B$ correlation stays flat because these methods do not use labels; LTR bias and clean-utility curves degrade as contamination rises.
  \item A controlled comparison against primary LTR peers (LightGBM, XGBoost, RankSVM) and secondary controls (coordinate ascent, linear weighted sum, Reciprocal Rank Fusion) on five listwise benchmarks.
  \item A reporting discipline: every bias claim is paired with clean-label $\ndcg$ on a bias--utility frontier.
\end{enumerate}

\subsection{Defining bias versus signal}
\label{sec:bias_def}
There is no single statistical test that separates ``bias'' from ``signal'' without a prior about what the system is for~\cite{blodgett2020language}. Closely related literature fix that prior differently: Geirhos et al.\ call \emph{shortcuts} the decision rules that succeed on a benchmark without capturing the intended concept, even when those rules remain predictive~\cite{geirhos2020shortcut}; ranking and recommender work treats over-reliance on popularity-like or structural correlates as bias relative to a stated goal~\cite{chen2023bias,abdollahpouri2019unfairness}; unbiased LTR treats clicks and presentation as \emph{biased feedback} when they systematically distort intended relevance~\cite{joachims2017unbiased,joachims2007evaluating}; and LTR practice further distinguishes the same behavioral quantity used as a contaminated \emph{label} versus as a ranking \emph{feature}~\cite{yang2022can}.

We use that same idea directly. The \emph{allowed signal} is the declared criteria vector~$C$; the clean label is used only to score utility. \emph{Bias} is dependence of the produced ranking on an undesignated field relative to that task definition: contaminant~$B$ (label-contamination study) or proxy~$P$ (proxy-skew study). Here~$P$ is a \emph{shortcut}~\cite{geirhos2020shortcut}: it can correlate with the clean label and still be outside the declared target~$C$ (e.g.\ review count or brand when~$C$ is rating, match, and price). Moderate label correlation makes the contrast sharp, relevance alone induces some rank--$P$ alignment, while locking onto~$P$ as a ranking feature pushes correlation well above that baseline.

Both~$B$ and~$P$ remain recorded for evaluation even when they are not inputs to a method. We report the rank--signal correlation of the produced ordering with~$B$ or~$P$ together with clean $\ndcg$. Under the same pools and metrics, Genetic~AI methods stay near the relevance-driven baseline for rank--$P$ correlation, while LightGBM, XGBoost, and RankSVM align with~$P$ far more strongly.

Table~\ref{tab:designations} states, for each benchmark, which raw field is undesignated and which attributes enter default~$C$. The label-contamination and proxy-skew studies use different datasets and different criteria vectors~$C$ even when the domain name matches; full proxy-skew weights appear in Appendix~\ref{app:criteria}.

\begin{table}[tb]
\centering
\caption{Per-benchmark designations. Undesignated~$B$/$P$ are refused by default~$C$; full proxy-skew weights/polarities are in Table~\ref{tab:criteria_pkg}.}
\label{tab:designations}
\scriptsize
\begin{tabularx}{\columnwidth}{@{}l>{\raggedright\arraybackslash}p{0.28\columnwidth}>{\raggedright\arraybackslash}X@{}}
\toprule
\textbf{Bench.} & \textbf{Undesignated} & \textbf{Default $C$ (summary)} \\
\midrule
Synthetic & Explicit constructed~$B$; $P$ vs.\ quality/price & quality, price (COST) \\
MSLR~\cite{qin2013introducing} & $B$: PageRank; $P$: \texttt{url\_clickcount} & Contam.: BM25/LM/TF/coverage (body). Proxy: BM25/LM/TF/stream length (body) \\
Baidu~\cite{zou2021pre} & $B$: BM25-mixed grades; $P$: doc.\ length & Contam.: BM25, QL-Dirichlet. Proxy: BM25 (title+body), QL variants, IDF \\
WANDS~\cite{chen2022wands} & \texttt{review\_count} as~$B$ and as~$P$ & Contam.: rating, size, height, class match. Proxy: class/name/desc/token match \\
Expedia~\cite{expedia2013} & $B$: inverted position; $P$: brand & Contam.\ / proxy: price, stars, review, location, destination distance (no position/brand) \\
\bottomrule
\end{tabularx}
\end{table}

\section{Related work}
\label{sec:related}

\subsection{Bias in learning to rank}
Click and presentation biases systematically distort relevance feedback~\cite{joachims2007evaluating}. Unbiased LTR corrects for position bias via inverse propensity scoring and related estimators~\cite{joachims2017unbiased,agarwal2019estimating,oosterhuis2023doubly}. Separately, popularity bias in recommenders and ranking amplifies head items~\cite{abdollahpouri2019unfairness,chen2023bias}. Our label-contamination study is closer to \emph{training-label} contamination than to online propensity correction: we mix a measurable contaminant~$B$ into grades and ask whether label-free ranking avoids absorbing~$B$.

\subsection{Societal bias and fair ranking}
Neural rankers can intensify gender and other societal biases relative to classical retrieval~\cite{rekabsaz2020neural,rekabsaz2021societal}. Fair ranking literature constrains exposure across providers or protected groups~\cite{singh2018fairness,geyik2019fairness,morik2020controlling,zehlike2022fair,ekstrand2022fairness}. We do not claim demographic fairness here; we study \emph{dependence on undesignated signals} (contaminants and proxies) that the task definition excludes from the optimization target. The methodological parallel is the insistence on trade-offs: fairness or bias metrics without utility are incomplete~\cite{ekstrand2022fairness,blodgett2020language}.

\subsection{Evolutionary and multi-criteria ranking}
Genetic programming has been used to discover information retrieval (IR) ranking functions~\cite{fan2004generic}. Fair ranking methods impose group-fairness constraints on ranked lists~\cite{tahery2021ga}. Classical evolutionary multi-objective optimization (e.g.\ {NSGA-II}) and evolutionary game theory provide the broader algorithmic background\cite{deb2002fast,smith1982evolution,holland1992adaptation}. A more recent line, Genetic~AI, casts discrete multi-objective decisions as evolutionary games over genes and organisms, without external preference weights or labelled training~\cite{wissgott2025geneticai}. Evolutionary Data Theory develops the corresponding formalism (replicator dynamics, evolutionary strategies, convergence) for matrix-form data problems~\cite{wissgott2026edt}, and related analysis establishes global convergence of feature-weighting dynamics on the simplex~\cite{daniilidis2026featureweighting}. We instantiate that paradigm for listwise reranking with the open-source \texttt{pikaia} package~\cite{wissgott2025geneticai}, comparing four Pikaia strategies: Dominant--Balanced (DOM-BAL: \texttt{DOMINANT} gene strategy with \texttt{BALANCED} organism strategy), REWARD-HARD, VALUATION-BLEND, and SELL-BUY. Unlike classical GP rankers that \emph{learn} a scoring formula from labels, these Genetic~AI methods optimizes the importance weights for attributes in a \emph{declared} criteria vector~$C$ at inference time on a fixed pool. Linear weighted sum and RRF~\cite{cormack2009reciprocal} serve as non-learned multi-criteria controls so that gains are not automatically attributed to the evolutionary search.

\subsection{Shortcut learning and proxies}
Supervised models often exploit spurious correlates of the training objective~\cite{geirhos2020shortcut}. The proxy-skew study demonstrates this phenomenon for LTR methods: a calibrated proxy~$P$ with moderate label correlation is available as a feature; we measure how much rankings behave like ``sort by~$P$.''

\section{Methodology}
\label{sec:method}

\subsection{Problem setting and evaluation splits}
Each evaluation unit is a \emph{query}: an information need paired with a fixed candidate \emph{pool} of roughly $20$--$100$ documents. The pool is never split into train/test documents; every method only reorders the same full pool for that query. Utility is always scored against \emph{clean} relevance labels using $\ndcg$~\cite{jarvelin2002cumulated}. Bias metrics measure dependence of the ranking on a proxy~$P$ (proxy-skew study) or an undesignated signal~$B$ (label-contamination study), as defined in Section~\ref{sec:bias_def}.

All train/test splits are at the \emph{query} level (disjoint sets of queries). Learned LTR methods are fit on the training queries and evaluated on held-out queries; Genetic~AI / linear / RRF do not fit on labels, they apply a fixed ranking rule to each held-out pool at evaluation time. How the criteria vector~$C$ is chosen is separate from this split and is detailed in Section~\ref{sec:criteria_protocol}.

\paragraph{Proxy-skew study (main claim).}
We reserve a \emph{selection} slice of $50$ queries and a disjoint \emph{evaluation} slice of the next $50$ queries. LTR is trained on the selection queries and scored on the evaluation queries. The same evaluation queries are used to score Genetic~AI and the other non-learned methods. The selection slice is also where~$C$ is declared (Section~\ref{sec:criteria_protocol}); reported metrics use only the evaluation slice.

\paragraph{Label-contamination study (supporting).}
We use a query-level train/test split of the sampled queries: LTR is trained on the train queries (with contaminated labels) and all methods are scored on the held-out queries.

\paragraph{Core reporting rule.}
Bias is never interpreted in isolation. A preferred method sits better on the frontier of clean $\ndcg$ (horizontal) vs.\ bias/skew (vertical): lower bias at matched utility, or higher utility at matched bias.

\subsection{Rankers and controls}
\label{sec:rankers}
Table~\ref{tab:methods} lists the shared registry. Primary learned peers for the paper claims are LightGBM, XGBoost, and RankSVM. Secondary controls are input order, sort-by-$B$/$P$, linear weighted sum on~$C$, RRF on~$C$, and coordinate ascent. Linear tests explicit criteria without the Genetic~AI search~\cite{holland1992adaptation,wissgott2025geneticai}.

\paragraph{Pikaia as a strategy framework.}
We treat Pikaia as an open Genetic~AI \emph{framework}, not a single fixed sorter~\cite{wissgott2025geneticai}. Candidates in a pool are organisms and declared criteria are genes. A run combines one or more gene-level strategies with one or more organism-level strategies, then runs the same iterative replicator updates for every strategy pair to obtain organism fitness scores used for ranking. All Genetic~AI algorithm arms below share the same criteria vector~$C$, preprocessing (pool-local scaling with COST polarity inverted), and evaluation protocol; only the strategy pair changes.

\paragraph{Strategies compared.}
\begin{itemize}[leftmargin=1.2em,itemsep=0.15em]
  \item \textbf{Pikaia (DOM-BAL)}: gene \texttt{DOMINANT} with organism \texttt{BALANCED} (Dominant--Balanced).
  \item \textbf{Pikaia (REWARD-HARD)}: gene \texttt{REWARD\_HARD} with \texttt{BALANCED}. Rewards rarely expressed (``hard'') criteria.
  \item \textbf{Pikaia (VALUATION-BLEND)}: gene\newline
    \texttt{VALUATION\_BLEND} ($\mathrm{preference}{=}0.3$) with \texttt{BALANCED}. Blends hard and easy valuation.
  \item \textbf{Pikaia (SELL-BUY)}: organism strategy \texttt{SELL} then \texttt{BUY} gene strategy.
\end{itemize}
Proxy ablations for the four Pikaia strategies are reported as diagnostics alongside the main results.

\begin{table}[tb]
\centering
\caption{Shared method registry.}
\label{tab:methods}
\footnotesize
\begin{tabularx}{\columnwidth}{@{}>{\raggedright\arraybackslash}p{0.48\columnwidth}X@{}}
\toprule
\textbf{Method} & \textbf{Role} \\
\midrule
Input order & Unsorted pool baseline \\
Sort by $B$ or $P$ & Upper bound on bias / skew \\
Linear weighted sum on $C$ & Explicit-criteria control \\
RRF on $C$~\cite{cormack2009reciprocal} & Non-learned multi-criteria peer \\
Pikaia (DOM-BAL / REWARD-HARD / VALUATION-BLEND / SELL-BUY)~\cite{wissgott2025geneticai,wissgott2026edt} & Open Genetic~AI framework, strategy variants \\
LightGBM~\cite{ke2017lightgbm,burges2010ranknet} & Primary tree listwise LTR peer \\
XGBoost~\cite{chen2016xgboost,burges2010ranknet} & Primary tree listwise LTR peer \\
RankSVM~\cite{joachims2002optimizing} & Primary pairwise LTR peer \\
Coord.\ ascent~\cite{metzler2007linear} & Secondary non-boosting learned control \\
\bottomrule
\end{tabularx}
\end{table}

\subsection{Proxy-skew protocol}
\label{sec:proxy_skew}
We use dedicated proxy-skew datasets with clean relevance labels (separate from the label-contamination datasets); designations of~$P$ and~$C$ are in Table~\ref{tab:designations}. All methods rerank the same pools under the shared evaluation protocol of Section~\ref{sec:method}. Learned LTR peers (LightGBM, XGBoost, RankSVM, coordinate ascent) are trained on clean labels; Genetic~AI algorithms, linear, and RRF apply their ranking rules at evaluation time. Ablations vary whether~$P$ enters the ranking inputs and are reported as $\Delta\mathrm{nDCG}$ (full minus no-$P$) together with rank--proxy correlation.

\paragraph{Keeping $C$ distinct from $P$.}
If attributes in~$C$ are nearly collinear with~$P$, then ranking by~$C$ alone can look like ranking by~$P$. We therefore choose structural or popularity proxies with moderate label correlation ($|\mathrm{corr}(P,\mathrm{label})|\approx 0.45$, target band $0.3$--$0.6$) and build a criteria vector~$C$ whose members have low pairwise correlation with~$P$.

\paragraph{How the criteria vector $C$ is declared.}
\label{sec:criteria_protocol}
Here~$C$ is the declared attribute vector used by Genetic~AI algorithms, linear, and RRF. We choose~$C$ once per domain on the selection slice in four steps, then keep that same~$C$ for evaluation. To choose how many attributes enter~$C$ without touching the evaluation slice, we further split the selection queries $80$/$20$: fit polarities and weights on $80\%$, and pick the criteria size~$k$ by $\mathrm{nDCG}$ on the held-out $20\%$. (i)~\emph{Proxy-hygiene gate}: we discard any candidate attribute with $|\mathrm{corr}(\text{attr},P)| > 0.35$ on the train split. (ii)~\emph{Polarity} (GAIN vs.\ COST): each attribute is treated as something to maximize or minimize. We default to a semantic guess from the name (e.g.\ rating $=$ GAIN, price $=$ COST) and override with the empirical train-split sign of $\mathrm{corr}(\text{attr},\text{label})$ only when that correlation has magnitude at least $0.05$. (iii)~\emph{Importance} (relative weight in~$C$): we set weights proportional to $|\mathrm{corr}(\text{attr},\text{label})|$ on the train split (monotone, no free parameters). (iv)~\emph{Criteria size}: we rank candidate attributes by importance, evaluate successive top-$k$ prefixes with the open \texttt{pikaia} engine on the internal validation split, and keep the~$k$ with the best validation $\mathrm{nDCG}$. As an additional candidate we also retain a hand-specified~$C$ written before the protocol (same attributes/polarities for that domain). On each domain the protocol picks whichever candidate wins on the internal validation split; on Expedia that hand-specified~$C$ wins, so the domain is unchanged.Table~\ref{tab:criteria_effect} records the evaluation-slice effect of switching from the manually selected~$C$ to the protocol-selected one. The selected proxy-skew~$C$ is frozen offline into the study configs (and listed in Appendix~\ref{app:criteria}) and reused unchanged on the evaluation slice.

\begin{table}[tb]
\centering
\caption{Effect of the criteria-selection protocol on the evaluation slice: hand-specified~$C$ $\rightarrow$ protocol-selected~$C$ (Pikaia). Expedia kept the hand-specified~$C$.}
\label{tab:criteria_effect}
\small
\begin{tabularx}{\columnwidth}{@{}lXX@{}}
\toprule
\textbf{Bench.} & $\Delta$nDCG & corr($P$) \\
\midrule
Baidu-PS & $+0.133$ & $0.45 \rightarrow 0.21$ \\
WANDS-PS & $+0.054$ & $0.17 \rightarrow 0.16$ \\
MSLR-PS & $+0.006$ & $0.13 \rightarrow 0.15$ \\
Expedia & $\pm 0.000$ & $0.20 \rightarrow 0.20$ \\
\bottomrule
\end{tabularx}
\end{table}

\subsection{Label-contamination protocol}
\label{sec:label_contamination}
We build identical candidate lists and mix the clean grade with min-max-normalized contaminant~$B$ at a stated contamination strength (Table~\ref{tab:designations}). We train LightGBM, XGBoost, RankSVM, and coordinate ascent on biased labels with a query-level train/test split. For Genetic~AI algorithms, linear, and RRF we use a declared criteria vector~$C$ of domain attributes that exclude~$B$; that criteria vector~$C$ is pre-declared per domain in the study and held fixed across all contamination strengths. We then rank with the four Pikaia strategies on that~$C$, linear($C$), RRF($C$), the biased LTR peers, sort-by-$B$, and input order, and score clean $\ndcg$ and dependence on~$B$. Point estimates use contamination $0.4$; sweeps over $[0,1]$ produce bias and utility curves.

Because Genetic~AI algorithms do not train on labels, their rankings cannot change with contamination. The useful empirical contrast is how LTR bias and clean utility change as contamination strength increases; the fixed-strength table is a supporting snapshot of those deltas.

\paragraph{How contamination is applied.}
Training labels in production ranking often mix intended relevance with presentation or popularity signals~\cite{joachims2007evaluating,joachims2017unbiased,abdollahpouri2019unfairness,chen2023bias}. For each query pool, let $y_i\in\{0,1,2\}$ be the clean relevance grade of document~$i$ (clamped to that range when the benchmark uses a wider scale) and let $b_i$ be the raw undesignated field from Table~\ref{tab:designations} (e.g.\ PageRank, review count, inverted position). We min--max normalize~$b_i$ within the pool to $\tilde{b}_i\in[0,1]$, then form a continuous mix at contamination strength $s\in[0,1]$,
\begin{equation}
  z_i(s)=(1-s)\,y_i + s\cdot(2\tilde{b}_i),
\end{equation}
and map $z_i(s)$ back to a training grade $\hat{y}_i(s)\in\{0,1,2\}$ by thresholds $1.5$ and $0.75$ (so $s{=}0$ recovers the clean grade and $s{=}1$ is fully contaminant-driven). LTR is trained on $\{\hat{y}_i(s)\}$; evaluation always uses the clean grades~$y_i$. Sweeps use $s\in\{0,0.2,0.4,0.6,0.8,1.0\}$; reported snapshots use $s{=}0.4$.

\subsection{Metrics}
\textbf{Utility (both studies):} $\ndcg$ on clean labels (primary), plus $\mathrm{MRR}$ / $\ndcg[5]$ as secondary diagnostics.

\textbf{Label-contamination study.}
\emph{Rank--contaminant correlation:} Spearman correlation between the produced ranking and contaminant~$B$ (higher means the ranking behaves more like sort-by-$B$).
\emph{Top-$k$ exposure lift:} fraction of high-$B$ items in the top-$k$ minus the pool base rate.
\emph{Bias amplification:} lift relative to the input-order baseline.
\emph{Bias slope ($\texttt{bias\_slope}$):} ordinary least-squares slope of rank--contaminant correlation against contamination strength over the sweep $[0,1]$ (a summary of the bias curve, not a utility score).

\textbf{Proxy-skew study.}
\emph{Rank--proxy correlation:} Spearman correlation between the produced ranking and proxy~$P$.
\emph{Exposure lift toward high-$P$ items:} top-$k$ high-$P$ fraction minus pool base rate.
\emph{Ablation $\Delta\mathrm{nDCG}$:} $\mathrm{nDCG}_{\mathrm{full}}-\mathrm{nDCG}_{\mathrm{no}\,P}$ (larger means more utility depends on~$P$).
\emph{Ablation $\Delta\mathrm{corr}$:} change in rank--proxy correlation under the same full vs.\ no-$P$ contrast.

\subsection{Benchmarks}
All benchmarks supply listwise candidate pools with per-item attributes; we declare the criteria vector~$C$ from those attributes, and take measurable~$B$ and/or~$P$ as designated in Table~\ref{tab:designations}. Fixed-strength label-contamination runs used $n{=}30$ (synthetic) or $n{=}50$ queries. Every proxy-skew number uses a disjoint evaluation slice of $n{=}50$ queries, separate from the $50$ used to select~$C$.

On the proxy-skew evaluation slice the realized mean $|\mathrm{corr}(P,\mathrm{label})|$ is $0.462$ (Expedia), $0.454$ (MSLR), and $0.405$ (Baidu), all inside the intended $0.3$--$0.6$ band, but only $0.252$ on WANDS (below the band). We report WANDS anyway and flag the shortfall: the proxy carries only weak label information on those queries, yet RankSVM's ranking still correlates $0.87$ with it. Read strictly, WANDS-PS is a weaker test of proxy \emph{calibration} and a stronger illustration of proxy \emph{over-reliance}.

\section{Results}
\label{sec:results}

\subsection{Hypotheses}
\textbf{Proxy-skew study.} LTR rank--proxy correlation exceeds Genetic~AI algorithms under multi-attribute~$C$. Utility is more sensitive to the proxy ablation for LTR than for Genetic~AI algorithms.

\textbf{Label-contamination study.} As contamination rises, LTR rank--contaminant correlation with~$B$ rises and clean $\mathrm{nDCG}$ tends to fall; Genetic~AI algorithms on fixed~$C$ stay stable on both axes. At matched clean utility, Genetic~AI algorithms show lower top-$k$ exposure of high-$B$ items than biased LTR.

\subsection{Proxy skew and ablation}
\label{sec:resB}

Table~\ref{tab:studyB} contrasts Genetic~AI algorithms with clean-label LTR on the primary domains; remaining datasets are in Appendix Table~\ref{tab:studyB_full}. Fig.~\ref{fig:B_proxy} plots proxy dependence and Table~\ref{tab:ablation} the ablation deltas. The corresponding WANDS-PS proxy--utility frontier appears as panel~(a) of the graphical abstract.

\begin{table}[tb]
\centering
\caption{Proxy-skew study, primary domains: Genetic~AI algorithms vs.\ clean-label LTR on the held-out query slice disjoint from criteria selection. Full benchmarks are in Table~\ref{tab:studyB_full}.}
\label{tab:studyB}
\small
\begin{tabularx}{\columnwidth}{@{}lXrr@{}}
\toprule
\textbf{Bench.} & \textbf{Method} & nDCG & corr($P$) \\
\midrule
\multirow{7}{*}{WANDS}
 & \textbf{Pikaia (DOM-BAL)} & $0.896$ & $0.164$ \\
 & \textbf{Pikaia (REWARD-HARD)} & $0.865$ & $0.147$ \\
 & \textbf{Pikaia (VALUATION-BLEND)} & $0.872$ & $0.151$ \\
 & \textbf{Pikaia (SELL-BUY)} & $0.886$ & $\mathbf{0.146}$ \\
 & LightGBM & $0.975$ & $0.586$ \\
 & XGBoost & $0.974$ & $0.785$ \\
 & RankSVM & $0.955$ & $0.867$ \\
\midrule
\multirow{7}{*}{MSLR}
 & \textbf{Pikaia (DOM-BAL)} & $0.350$ & $0.148$ \\
 & \textbf{Pikaia (REWARD-HARD)} & $0.350$ & $0.138$ \\
 & \textbf{Pikaia (VALUATION-BLEND)} & $0.343$ & $0.144$ \\
 & \textbf{Pikaia (SELL-BUY)} & $0.320$ & $\mathbf{0.137}$ \\
 & LightGBM & $0.965$ & $0.428$ \\
 & XGBoost & $0.941$ & $0.442$ \\
 & RankSVM & $0.762$ & $0.462$ \\
\bottomrule
\end{tabularx}

\end{table}

\begin{figure}[tb]
\centering
\includegraphics[width=\columnwidth]{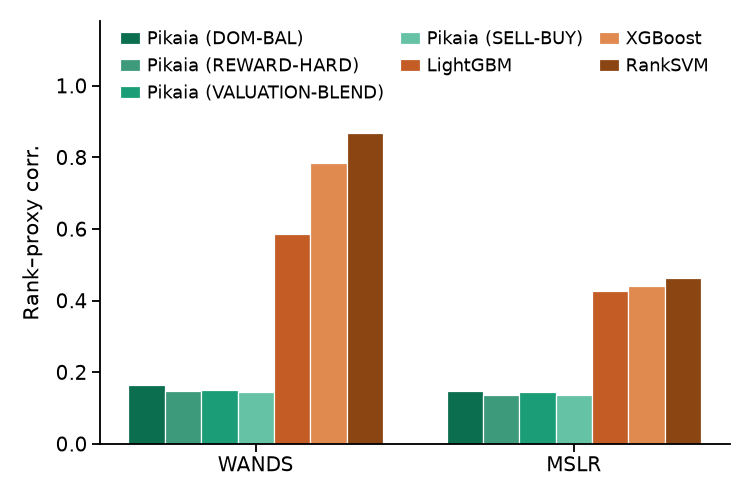}
\caption{Proxy-skew study: proxy dependence (WANDS, MSLR). Genetic~AI algorithms show lower proxy dependence than clean-label LTR.}
\label{fig:B_proxy}
\end{figure}

\textbf{Headline findings.} Clean-label LTR peers show higher rank--proxy correlation than Genetic~AI algorithms. On WANDS-PS, proxy corr is $0.59$/$0.78$/$0.87$ (LightGBM, XGBoost, RankSVM) vs.\ Genetic~AI algorithms $\approx 0.15$--$0.16$. The same ordering holds on MSLR-PS.On the graphical-abstract frontier (inverted $y$-axis), Genetic~AI algorithms keep the preferred upper-right corner but concede clean $\mathrm{nDCG}$ to LightGBM, XGBoost, and RankSVM.

\textbf{Ablation.} Removing~$P$ hurts LightGBM more than Pikaia on MSLR ($0.370$ vs.\ $0.123$) and Baidu ($0.162$ vs.\ $0.114$); on WANDS the two are effectively tied ($0.022$ vs.\ $0.024$), because Genetic~AI algorithms are already near ceiling. Expedia reverses the ordering outright ($0.128$ vs.\ $0.316$), so we do not read the pattern as universal: Expedia's criteria set is the narrowest of the four, so a forced proxy gene supplies proportionally more of the signal there than it does elsewhere. Pikaia proxy ablations likewise show large utility deltas on some domains. Linear tracks Genetic~AI algorithms closely, which again points at explicit criteria rather than evolutionary search as the active ingredient. Lower proxy alignment for Genetic~AI holds on all four domains; the stronger utility-ablation pattern holds on two.

\begin{table}[tb]
\centering
\caption{Ablation $\Delta\mathrm{nDCG}=\mathrm{nDCG}_{\mathrm{full}}-\mathrm{nDCG}_{\mathrm{no}\,P}$. Larger means more utility depends on~$P$.}
\label{tab:ablation}
\small
\begin{tabularx}{\columnwidth}{@{}lXXXX@{}}
\toprule
\textbf{Family} & \textbf{Exp.} & \textbf{MSLR} & \textbf{WANDS} &
\textbf{Baidu} \\
\midrule
LightGBM & $0.128$ & $0.370$ & $0.022$ & $0.162$ \\
\textbf{Pikaia} & $0.316$ & $0.123$ & $0.024$ & $0.114$ \\
linear & $0.462$ & $0.131$ & $0.038$ & $0.160$ \\
\bottomrule
\end{tabularx}
\end{table}

\subsection{Label contamination}
\label{sec:resA}

\begin{figure}[tb]
\centering
\includegraphics[width=\columnwidth]{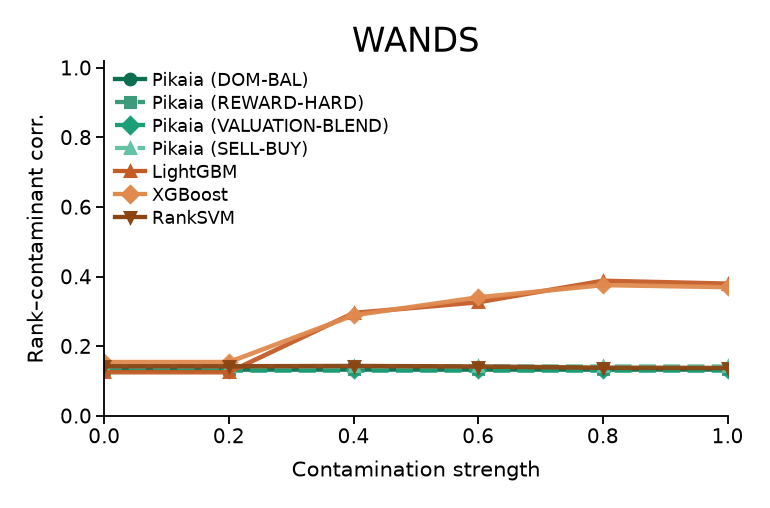}
\caption{Label-contamination study: contamination immunity on WANDS (Baidu shows the same flat Genetic~AI pattern; Table~\ref{tab:studyA}). Genetic~AI algorithms stay flat; LTR absorbs the contaminant.}
\label{fig:A_immunity}
\end{figure}

\begin{figure}[tb]
\centering
\includegraphics[width=\columnwidth]{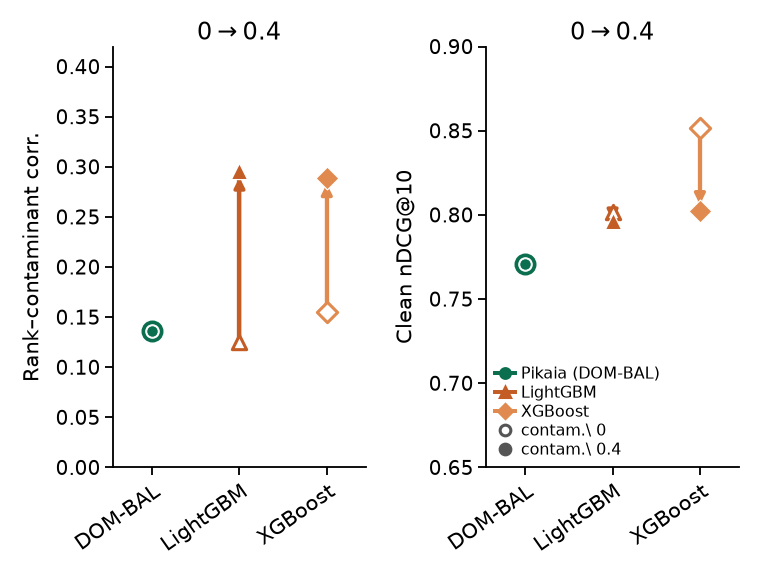}
\caption{Label-contamination study: before/after change on WANDS from contamination~$0$ to~$0.4$. Left: rank--contaminant correlation; right: clean $\ndcg$. Open markers at~$0$, filled at~$0.4$; Genetic~AI (DOM-BAL) is flat on both axes while LightGBM/XGBoost rise in bias (and lose some utility).}
\label{fig:A_before_after}
\end{figure}

As a supporting check, the label-contamination study confirms that Genetic~AI algorithm rankings are invariant to contaminated training grades (they never train on labels) and shows how LTR bias and clean utility \emph{change} when~$B$ is mixed into labels. Figures~\ref{fig:A_immunity} and~\ref{fig:A_before_after} lead with the relative claim (continuous sweep; before/after deltas on WANDS); Table~\ref{tab:studyA} is a matched-strength snapshot at contamination~$0.4$. Full benchmarks are in Appendix~\ref{app:tables}.

\begin{table}[tb]
\centering
\caption{Label-contamination study, primary domains (contamination $0.4$ snapshot). Higher $\ndcg$ is better; lower rank-corr / top-$k$ lift are better on bias.}
\label{tab:studyA}
\small
\begin{tabularx}{\columnwidth}{@{}lXrrr@{}}
\toprule
\textbf{Bench.} & \textbf{Method} & nDCG & corr & lift \\
\midrule
\multirow{7}{*}{Baidu}
 & \textbf{Pikaia (DOM-BAL)} & $0.424$ & $0.188$ & $-0.066$ \\
 & \textbf{Pikaia (REWARD-HARD)} & $0.422$ & $0.185$ & $-0.066$ \\
 & \textbf{Pikaia (VALUATION-BLEND)} & $0.419$ & $0.182$ & $-0.072$ \\
 & \textbf{Pikaia (SELL-BUY)} & $0.427$ & $0.188$ & $\mathbf{-0.078}$ \\
 & LightGBM & $0.446$ & $\mathbf{0.176}$ & $-0.006$ \\
 & XGBoost & $0.486$ & $0.179$ & $-0.026$ \\
 & RankSVM & $0.374$ & $0.375$ & $0.176$ \\
\midrule
\multirow{7}{*}{WANDS}
 & \textbf{Pikaia (DOM-BAL)} & $0.771$ & $0.136$ & $-0.020$ \\
 & \textbf{Pikaia (REWARD-HARD)} & $0.732$ & $0.135$ & $-0.004$ \\
 & \textbf{Pikaia (VALUATION-BLEND)} & $0.764$ & $\mathbf{0.131}$ & $-0.026$ \\
 & \textbf{Pikaia (SELL-BUY)} & $0.718$ & $0.145$ & $-0.014$ \\
 & LightGBM & $0.796$ & $0.296$ & $0.036$ \\
 & XGBoost & $0.803$ & $0.289$ & $0.034$ \\
 & RankSVM & $0.757$ & $0.143$ & $\mathbf{-0.038}$ \\
\bottomrule
\end{tabularx}

\end{table}

Relative to clean training labels, LightGBM's WANDS rank--$B$ correlation rises from $0.126$ at contamination~$0$ to $0.296$ at~$0.4$ and $0.380$ at~$1.0$; XGBoost moves from $0.155$ to $0.289$ / $0.370$ on the same grid (Fig.~\ref{fig:A_before_after}). Clean $\ndcg$ for LightGBM falls from $0.802$ at~$0$ to $0.797$ at~$0.4$ and $0.716$ at~$1.0$ (XGBoost $0.851 \rightarrow 0.803 \rightarrow 0.714$). Genetic~AI algorithm rank--$B$ correlation and clean $\ndcg$ stay flat across $[0,1]$ (Fig.~\ref{fig:A_immunity}; DOM-BAL corr.\ $0.136$, $\ndcg$~$0.771$). OLS bias slopes summarize the curves ($\texttt{bias\_slope}\approx 0.58$--$0.65$ for LightGBM/XGBoost on WANDS/Baidu). At the $0.4$ snapshot (Table~\ref{tab:studyA}), Genetic~AI algorithms nearly match biased tree utility on Baidu/WANDS while cutting contaminant correlation roughly in half. Expedia remains the honest trade-off (Appendix Table~\ref{tab:studyA_full}).

\section{Discussion}
\label{sec:discussion}

The main claim is that genetic multi-criteria sorting on a declared criteria vector~$C$ occupies a more favourable bias--utility operating point than clean-label LTR: substantially lower rank--proxy correlation at competitive clean $\mathrm{nDCG}$. On WANDS/MSLR in the proxy-skew study, Pikaia strategies stay far less proxy-aligned than LightGBM, XGBoost, and RankSVM while remaining competitive on clean $\mathrm{nDCG}$. Supporting label-contamination evidence from WANDS/Baidu shows flat contamination curves for Genetic~AI algorithms and rising bias (with falling clean utility) for label-trained LTR. Weaker domains (Expedia utility gap; large proxy-skew $\mathrm{nDCG}$ gaps vs.\ full LTR) are reported in the appendix without being elevated to the main claim.

\paragraph{Caveats.}
(i)~When true quality is entangled with~$B$ (Expedia position), LTR can win raw $\mathrm{nDCG}$; the question is whether that utility is \emph{allowed}. (ii)~Linear weighted sum often sits near Genetic~AI algorithms; Genetic~AI algorithms add value when non-linear multi-criteria trade-offs beat a fixed linear form. (iii)~We do not claim Genetic~AI algorithms always win on $\mathrm{nDCG}$; we claim they occupy a better robustness operating point when contaminants or proxies are excluded from the optimization target. (iv)~Query counts are moderate ($n{\approx}30$--$50$); results should be read as controlled evidence, not large-scale online evaluation. (v)~Societal protected-attribute fairness~\cite{rekabsaz2021societal,zehlike2022fair} is out of scope.

\paragraph{Implications.}
When the goal is to optimize only the declared attributes~$C$ and avoid popularity / position / brand shortcuts, genetic multi-criteria sorting does exactly that, with flat contamination response and low proxy alignment---properties label-trained LTR does not share even when labels are clean. Hybrid pipelines (LTR for recall, Genetic~AI algorithms for constrained re-ranking) are a natural next step.

\section{Conclusion}
\label{sec:conclusion}

We compared genetic multi-criteria sorting to label-trained LTR under two failure modes (contaminated training labels and clean-label proxy skew) across five listwise benchmarks. The main result is from the proxy-skew study: default criteria vectors remain far less proxy-aligned than LightGBM, XGBoost, or RankSVM on a bias--utility frontier. Supporting label-contamination evidence shows contamination sweeps leave Genetic~AI algorithms unaffected on the bias axis while LTR bias curves rise and clean utility falls. The appropriate claim is robustness under that refusal, not universal dominance on $\mathrm{nDCG}$. Future work includes larger query samples, online evaluation, protected-attribute fairness metrics, and hybrid LTR+Genetic~AI pipelines.

\section*{Data and code availability}
Experiment protocols, criteria configs, and evaluation scripts are maintained in the accompanying research repository; proxy-skew criteria vectors are also listed in Appendix~\ref{app:criteria}. Benchmark datasets are used under their respective licenses (MSLR-WEB10K, Baidu-ULTR, WANDS, Expedia/ICDM).

\section*{Author contributions}
Vladyslav Shevchuk: Conceptualization, Methodology, Software, Validation, Formal analysis, Investigation, Data curation, Writing --- review \& editing, Visualization.
Carlos Uziel P\'erez Malla: Supervision, Writing --- review \& editing.
Philipp Wissgott: Conceptualization, Supervision, Writing --- review \& editing.

\bibliography{references}

\clearpage
\appendix
\onecolumngrid
\raggedbottom
\setlength{\parskip}{0.4em plus 0.1em minus 0.1em}
\renewcommand{\arraystretch}{1.15}

\section{Full result tables}
\label{app:tables}

Tables~\ref{tab:studyA_full} and~\ref{tab:studyB_full} report every benchmark we
ran, including the domains where Genetic~AI sorting pays a visible utility price.
They are given here in full so that the primary-domain summaries in
Section~\ref{sec:results} can be checked against the complete picture.

\medskip
\noindent\refstepcounter{table}\label{tab:studyA_full}%
\textbf{\tablename~\thetable.}~Label-contamination study, all benchmarks at contamination $0.4$. Higher nDCG is better; lower rank-corr and top-$k$ lift are better.\\[0.5em]
{\scriptsize\centering
\begin{tabularx}{\textwidth}{@{}lXrrr@{}}
\toprule
\textbf{Bench.} & \textbf{Method} & nDCG & corr & lift \\
\midrule
\multirow{7}{*}{Baidu}
 & \textbf{Pikaia (DOM-BAL)} & $0.424$ & $0.188$ & $-0.066$ \\
 & \textbf{Pikaia (REWARD-HARD)} & $0.422$ & $0.185$ & $-0.066$ \\
 & \textbf{Pikaia (VALUATION-BLEND)} & $0.419$ & $0.182$ & $-0.072$ \\
 & \textbf{Pikaia (SELL-BUY)} & $0.427$ & $0.188$ & $\mathbf{-0.078}$ \\
 & LightGBM & $0.446$ & $\mathbf{0.176}$ & $-0.006$ \\
 & XGBoost & $0.486$ & $0.179$ & $-0.026$ \\
 & RankSVM & $0.374$ & $0.375$ & $0.176$ \\
\midrule
\multirow{7}{*}{WANDS}
 & \textbf{Pikaia (DOM-BAL)} & $0.771$ & $0.136$ & $-0.020$ \\
 & \textbf{Pikaia (REWARD-HARD)} & $0.732$ & $0.135$ & $-0.004$ \\
 & \textbf{Pikaia (VALUATION-BLEND)} & $0.764$ & $\mathbf{0.131}$ & $-0.026$ \\
 & \textbf{Pikaia (SELL-BUY)} & $0.718$ & $0.145$ & $-0.014$ \\
 & LightGBM & $0.796$ & $0.296$ & $0.036$ \\
 & XGBoost & $0.803$ & $0.289$ & $0.034$ \\
 & RankSVM & $0.757$ & $0.143$ & $\mathbf{-0.038}$ \\
\midrule
\multirow{7}{*}{Expedia}
 & \textbf{Pikaia (DOM-BAL)} & $0.354$ & $0.334$ & $0.050$ \\
 & \textbf{Pikaia (REWARD-HARD)} & $0.306$ & $0.296$ & $0.046$ \\
 & \textbf{Pikaia (VALUATION-BLEND)} & $0.349$ & $0.296$ & $0.056$ \\
 & \textbf{Pikaia (SELL-BUY)} & $0.220$ & $\mathbf{0.255}$ & $\mathbf{0.010}$ \\
 & LightGBM & $0.667$ & $0.376$ & $0.144$ \\
 & XGBoost & $0.654$ & $0.349$ & $0.138$ \\
 & RankSVM & $0.327$ & $0.306$ & $0.100$ \\
\midrule
\multirow{7}{*}{MSLR}
 & \textbf{Pikaia (DOM-BAL)} & $0.397$ & $0.127$ & $-0.013$ \\
 & \textbf{Pikaia (REWARD-HARD)} & $0.390$ & $0.120$ & $-0.007$ \\
 & \textbf{Pikaia (VALUATION-BLEND)} & $0.390$ & $0.129$ & $-0.011$ \\
 & \textbf{Pikaia (SELL-BUY)} & $0.377$ & $0.131$ & $\mathbf{-0.023}$ \\
 & LightGBM & $0.448$ & $0.104$ & $0.013$ \\
 & XGBoost & $0.492$ & $0.109$ & $0.013$ \\
 & RankSVM & $0.380$ & $\mathbf{0.097}$ & $-0.019$ \\
\midrule
\multirow{7}{*}{Synth.}
 & \textbf{Pikaia (DOM-BAL)} & $0.910$ & $0.340$ & $\mathbf{0.000}$ \\
 & \textbf{Pikaia (REWARD-HARD)} & $0.877$ & $0.344$ & $\mathbf{0.000}$ \\
 & \textbf{Pikaia (VALUATION-BLEND)} & $0.872$ & $0.390$ & $\mathbf{0.000}$ \\
 & \textbf{Pikaia (SELL-BUY)} & $0.909$ & $\mathbf{0.248}$ & $\mathbf{0.000}$ \\
 & LightGBM & $0.996$ & $0.321$ & $\mathbf{0.000}$ \\
 & XGBoost & $0.976$ & $0.313$ & $\mathbf{0.000}$ \\
 & RankSVM & $0.932$ & $0.282$ & $\mathbf{0.000}$ \\
\bottomrule
\end{tabularx}
\par}

\newpage
\noindent\refstepcounter{table}\label{tab:studyB_full}%
\textbf{\tablename~\thetable.}~Proxy-skew study, all benchmarks: Genetic~AI algorithms vs.\ clean-label LTR.\\[0.5em]
{\scriptsize\centering
\begin{tabularx}{\textwidth}{@{}lXrr@{}}
\toprule
\textbf{Bench.} & \textbf{Method} & nDCG & corr($P$) \\
\midrule
\multirow{7}{*}{Expedia}
 & \textbf{Pikaia (DOM-BAL)} & $0.319$ & $0.197$ \\
 & \textbf{Pikaia (REWARD-HARD)} & $0.309$ & $\mathbf{0.188}$ \\
 & \textbf{Pikaia (VALUATION-BLEND)} & $0.315$ & $0.203$ \\
 & \textbf{Pikaia (SELL-BUY)} & $0.251$ & $0.232$ \\
 & LightGBM & $0.990$ & $0.485$ \\
 & XGBoost & $0.993$ & $0.502$ \\
 & RankSVM & $0.946$ & $0.727$ \\
\midrule
\multirow{7}{*}{MSLR-PS}
 & \textbf{Pikaia (DOM-BAL)} & $0.350$ & $0.148$ \\
 & \textbf{Pikaia (REWARD-HARD)} & $0.350$ & $0.138$ \\
 & \textbf{Pikaia (VALUATION-BLEND)} & $0.343$ & $0.144$ \\
 & \textbf{Pikaia (SELL-BUY)} & $0.320$ & $\mathbf{0.137}$ \\
 & LightGBM & $0.965$ & $0.428$ \\
 & XGBoost & $0.941$ & $0.442$ \\
 & RankSVM & $0.762$ & $0.462$ \\
\midrule
\multirow{7}{*}{WANDS-PS}
 & \textbf{Pikaia (DOM-BAL)} & $0.896$ & $0.164$ \\
 & \textbf{Pikaia (REWARD-HARD)} & $0.865$ & $0.147$ \\
 & \textbf{Pikaia (VALUATION-BLEND)} & $0.872$ & $0.151$ \\
 & \textbf{Pikaia (SELL-BUY)} & $0.886$ & $\mathbf{0.146}$ \\
 & LightGBM & $0.975$ & $0.586$ \\
 & XGBoost & $0.974$ & $0.785$ \\
 & RankSVM & $0.955$ & $0.867$ \\
\midrule
\multirow{7}{*}{Baidu-PS}
 & \textbf{Pikaia (DOM-BAL)} & $0.437$ & $0.214$ \\
 & \textbf{Pikaia (REWARD-HARD)} & $0.444$ & $0.230$ \\
 & \textbf{Pikaia (VALUATION-BLEND)} & $0.427$ & $\mathbf{0.204}$ \\
 & \textbf{Pikaia (SELL-BUY)} & $0.442$ & $0.217$ \\
 & LightGBM & $0.812$ & $0.697$ \\
 & XGBoost & $0.800$ & $0.659$ \\
 & RankSVM & $0.704$ & $0.651$ \\
\bottomrule
\end{tabularx}
\par}

\newpage
\section{Declared criteria vectors for the proxy-skew study}
\label{app:criteria}

Table~\ref{tab:criteria_pkg} lists the criteria vector~$C$ used by every
non-LTR method on each proxy-skew domain, as selected by the train-split protocol
of Section~\ref{sec:criteria_protocol}. No vector contains~$P$. Weights are
normalized to sum to one and polarity is GAIN unless marked.

One polarity is worth flagging: WANDS \texttt{class\_}\allowbreak\texttt{token\_}\allowbreak\texttt{match} is selected
as a COST. Semantic inference proposes GAIN, but the train-split correlation
with the clean grade is negative with magnitude above the override threshold,
because the attribute fires on broad category words that match many
off-intent products in the same class. We report the selected direction rather
than overriding it by hand, and note that it carries the smallest weight in
that vector.

\medskip
\noindent\refstepcounter{table}\label{tab:criteria_pkg}%
\textbf{\tablename~\thetable.}~Proxy-skew criteria vectors selected on the train split. Polarity is GAIN unless marked COST.\\[0.5em]
{\scriptsize\centering
\begin{tabularx}{\textwidth}{@{}lXr@{}}
\toprule
\textbf{Bench.} & \textbf{Attribute} & \textbf{Weight} \\
\midrule
\multirow{4}{*}{WANDS-PS}
 & \texttt{product\_class\_match} & $0.364$ \\
 & \texttt{name\_match} & $0.253$ \\
 & \texttt{desc\_match} & $0.241$ \\
 & \texttt{class\_token\_match} (COST) & $0.142$ \\
\midrule
\multirow{4}{*}{MSLR-PS}
 & \texttt{bm25\_body} & $0.289$ \\
 & \texttt{lm\_dir\_body} & $0.260$ \\
 & \texttt{sum\_tf\_body} & $0.254$ \\
 & \texttt{stream\_length\_body} & $0.198$ \\
\midrule
\multirow{5}{*}{Baidu-PS}
 & \texttt{bm25\_title} & $0.216$ \\
 & \texttt{bm25} & $0.205$ \\
 & \texttt{ql\_jelinek\_mercer\_long} & $0.196$ \\
 & \texttt{ql\_dirichlet} & $0.193$ \\
 & \texttt{idf} & $0.191$ \\
\midrule
\multirow{5}{*}{Expedia}
 & \texttt{price\_usd} (COST) & $0.25$ \\
 & \texttt{prop\_starrating} & $0.20$ \\
 & \texttt{prop\_review\_score} & $0.20$ \\
 & \texttt{prop\_location\_score2} & $0.20$ \\
 & \texttt{orig\_destination\_distance} (COST) & $0.15$ \\
\bottomrule
\end{tabularx}
\par}

\end{document}